\documentclass{egpubl}
\usepackage{sca2020}

\SpecialIssuePaper       

\usepackage[T1]{fontenc}
\usepackage{dfadobe}  

\usepackage{cite}  
\BibtexOrBiblatex
\electronicVersion
\PrintedOrElectronic
\ifpdf \usepackage[pdftex]{graphicx} \pdfcompresslevel=9
\else \usepackage[dvips]{graphicx} \fi

\usepackage{egweblnk} 

\title{ZooBuilder: 2D and 3D Pose Estimation for Quadrupeds Using Synthetic Data
}
\author[A.S. Fangbemi \& Y.F. Lu \& M.Y Xu \& X.W. Luo \& A. Rolland \& C. Raissi]
{\parbox{\textwidth}{\centering A.\,S. Fangbemi$^1$, Y.\,F. Lu$^1$, M.\,Y. Xu$^2$, X.\,W. Luo$^1$, A. Rolland$^1$, C. Raissi$^{3,4}$}
\\
{\parbox{\textwidth}{\centering $^1$Ubisoft China AI \& Data Lab\\
         $^2$Sichuan University\\
         $^3$Ubisoft, Singapore\\
         $^4$INRIA Nancy Grand Est, France}
}
\vspace*{-1cm}
}
\usepackage{mathtools}

\usepackage{titlesec}

\begin{document}

\teaser{
 \includegraphics[width=\linewidth]{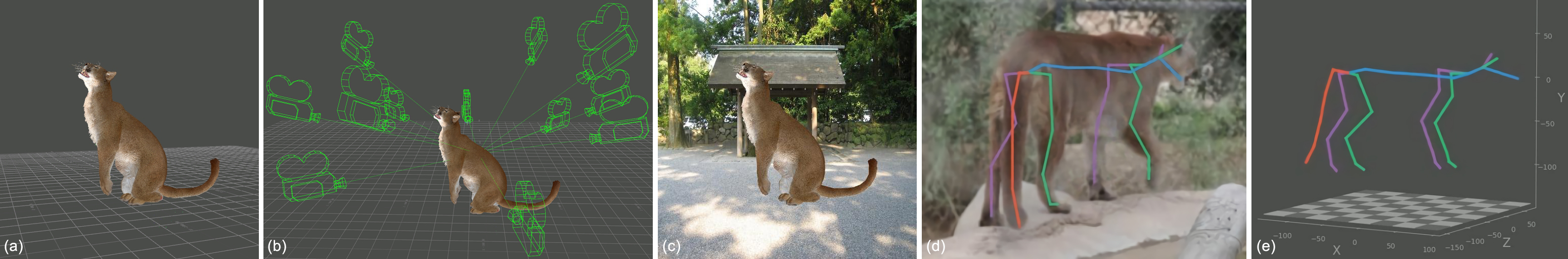}
 \centering
 \caption{Process from data generation to 3D pose estimation: (a) Original FBX animation file, (b) Camera setup in MotionBuilder, (c) sample training data, (d) 2D pose inference, (e) 3D pose inference.}
\label{fig:teaser}
}

\maketitle
\begin{abstract}
This work introduces a novel strategy for generating synthetic training data for 2D and 3D pose estimation of animals using keyframe animations. With the objective to automate the process of creating animations for wildlife, we train several 2D and 3D pose estimation models with synthetic data, and put in place an end-to-end pipeline called \textbf{ZooBuilder}. The pipeline takes as input a video of an animal in the wild, and generates the corresponding 2D and 3D coordinates for each joint of the animal's skeleton. With this approach, we produce motion capture data that can be used to create animations for wildlife.


\end{abstract}


\titlespacing\section{0pt}{4pt plus 2pt minus 2pt}{0pt plus 2pt minus 2pt}
\titlespacing\subsection{0pt}{4pt plus 2pt minus 2pt}{0pt plus 2pt minus 2pt}
\titlespacing\subsubsection{0pt}{4pt plus 2pt minus 2pt}{0pt plus 2pt minus 2pt}
\section{Introduction}
Thanks to the relative abundance of training datasets for humans, researchers have developed reliable deep learning approaches for 2D and 3D pose estimation\cite{cao2018openpose}. These datasets are usually created in a controlled environment such as a motion capture room where actors perform a variety of actions. In contrast to humans, bringing wild animals in a mocap room can be a challenging task which limits the availability of training datasets and research on the topic. In this paper, we address the difficult work of generating animal datasets for quadrupeds pose estimation by creating synthetic data from existing keyframe animations. Given an FBX animation file of a cougar, we create a set of cameras around it to build a virtual mocap room (Figure \ref{fig:teaser}). From each camera view and for each frame of the animation, we extract the 2D and 3D coordinates of the skeleton, an image, as well as the parameters of the camera. Using this data, we retrain 2D and a 3D pose estimation models originally developed for humans. We include these models in an end-to-end pipeline called \emph{“ZooBuilder”}, which takes as input a video of a cougar and outputs the corresponding 3D animation.

\section{Approach}
\paragraph*{Training data generation.}
\begin{figure}[htb]
  \centering
  \includegraphics[width=1\linewidth]{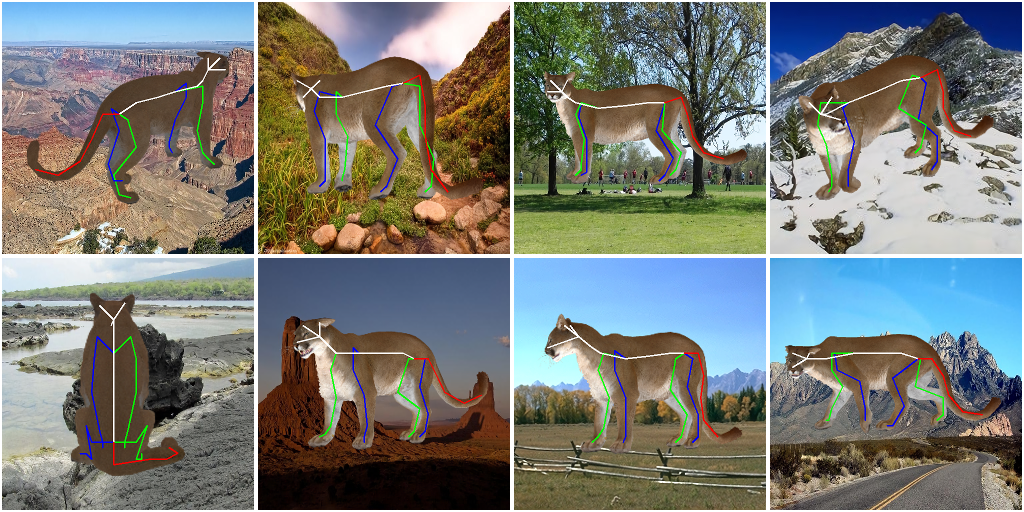}
  \caption{\label{fig:zbDataset}
           Samples of the dataset used for training 2D pose estimation models.}
\end{figure}

We import FBX files containing keyframe animations of a cougar into Autodesk MotionBuilder and create a set of twelve cameras. For each camera and frame, we generate: (1) the 3D world coordinates for $37$ joints, (2) a rendered image, (3) the 3D world coordinates of the camera, its projection matrix and the 3D world coordinates of its focal point. We further generate the 3D view coordinates of the joints from the camera perspective by multiplying their 3D world coordinates by the view matrix. The 3D view coordinates are then used to compute the 2D coordinates of the joints on the projection plane, which are finally converted to 2D image coordinates.\cite{dunn20103d}.

In addition, we post process images generated from MotionBuilder to add random backgrounds in order to increase their realism \cite{zhou2016places}. The final training dataset consists of (1) cougar images with background, (2) 2D image coordinates (Figure \ref{fig:zbDataset}), and (3) 3D view coordinates of each joint for each image.

\paragraph*{ZooBuilder pipeline.}
ZooBuilder is an end-to-end pipeline that generates FBX animation files from 2D videos of animals. It involves three core steps leveraging machine learning models (ML) for (1) object detection (YoLoV3\cite{redmon2018yolov3}), (2) 2D pose estimation (OpenPose\cite{cao2018openpose}), and (3) 3D pose estimation (Pose\_3D\cite{rayat2018exploiting}). We adapt the 2D and 3D pose estimation models to our synthetic cougar data and retrain them.

\section{Challenges}
\begin{figure}[htb]
  \centering
  \includegraphics[width=1\linewidth]{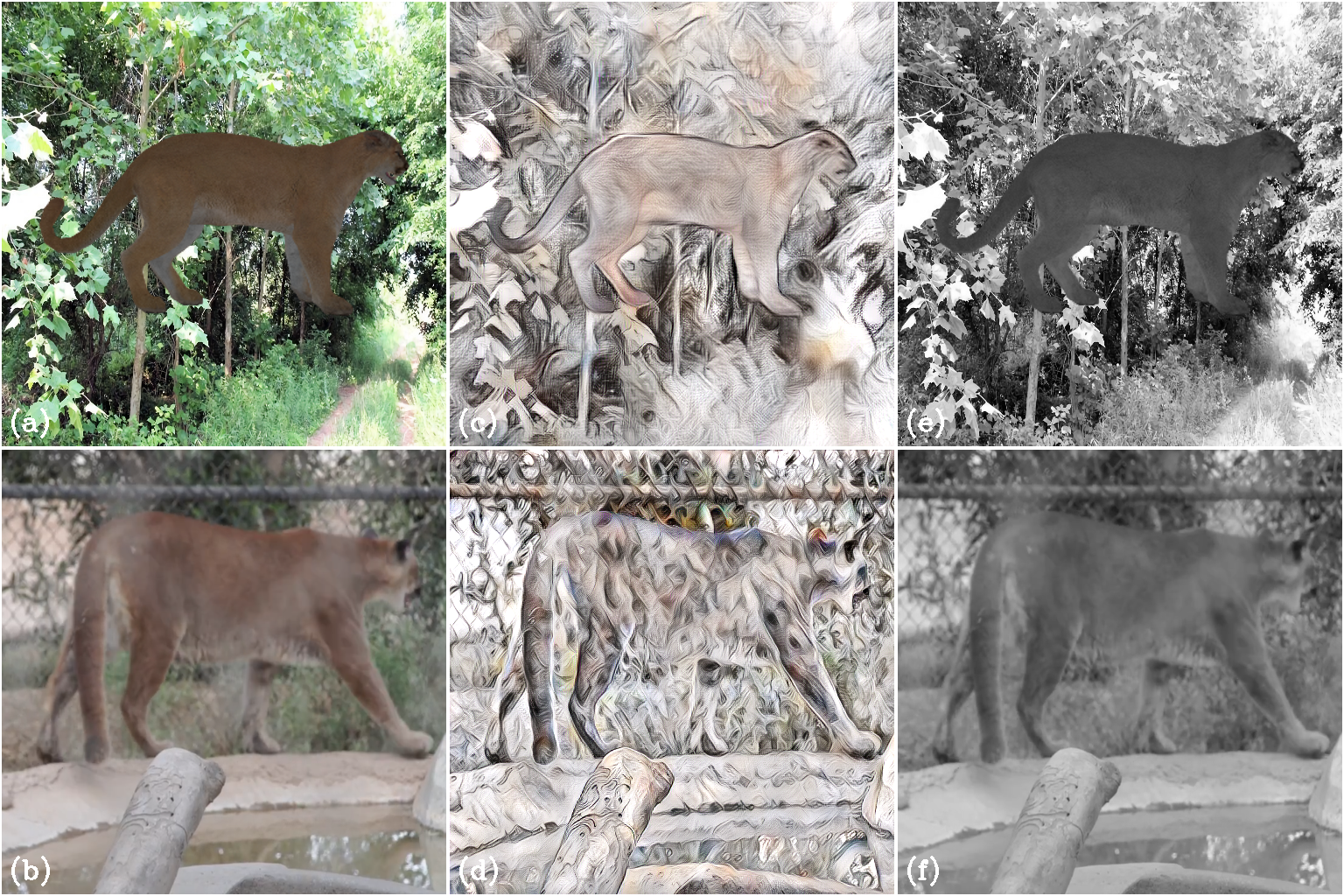}
  \caption{\label{fig:zbDataTrans}
           Transformation to standardize train data (above) and test data (below) with style transfer (c, d) or gray-scale(e, f). }
    \vspace{-3mm}
\end{figure}

\paragraph*{Irrealistic training data.} The main challenge to overcome with our approach resides in the difference of distribution between the synthetic training data and the real test data collected from the wild. To ensure that models trained on synthetic data generalize well on realistic videos, we explore two main data transformation techniques, changing both training and test images into (1) gray-scale images or (2) using a style transfer technique to standardize the data (Figure \ref{fig:zbDataTrans}). Although training and testing our 2D pose estimation models on gray-scale images results in a significant improvement in the prediction accuracy, the results with style transfer under perform predictions done without style transfer. 

\paragraph*{Training data volume.}
ML models are usually trained with a large quantity of images. Using transfer learning strategies, we leverage features learned by Resnet50 on the ImageNet dataset, and further train our 2D pose estimation model with our synthetic data composed of about $170$ thousand images. To increase the diversity of images, we introduce different data augmentation techniques during the training phase such as image and joints rotation, scaling, flipping, Gaussian noise, color jittering, random brightness and contrast.

\paragraph*{Machine learning models tuning.}
Apart from the challenges mentioned above, retraining human based models for a cougar skeleton and dataset requires to invest efforts in fine-tuning the models. We explore different hyper-parameters configurations for the learning rate, optimizer, activation functions. We adopt a layer-specific learning rate approach coupled with a learning scheduler with the SGD optimizer to train a model that generalizes relatively well.



\begin{figure}[htb]
  \centering
  \includegraphics[width=1\linewidth]{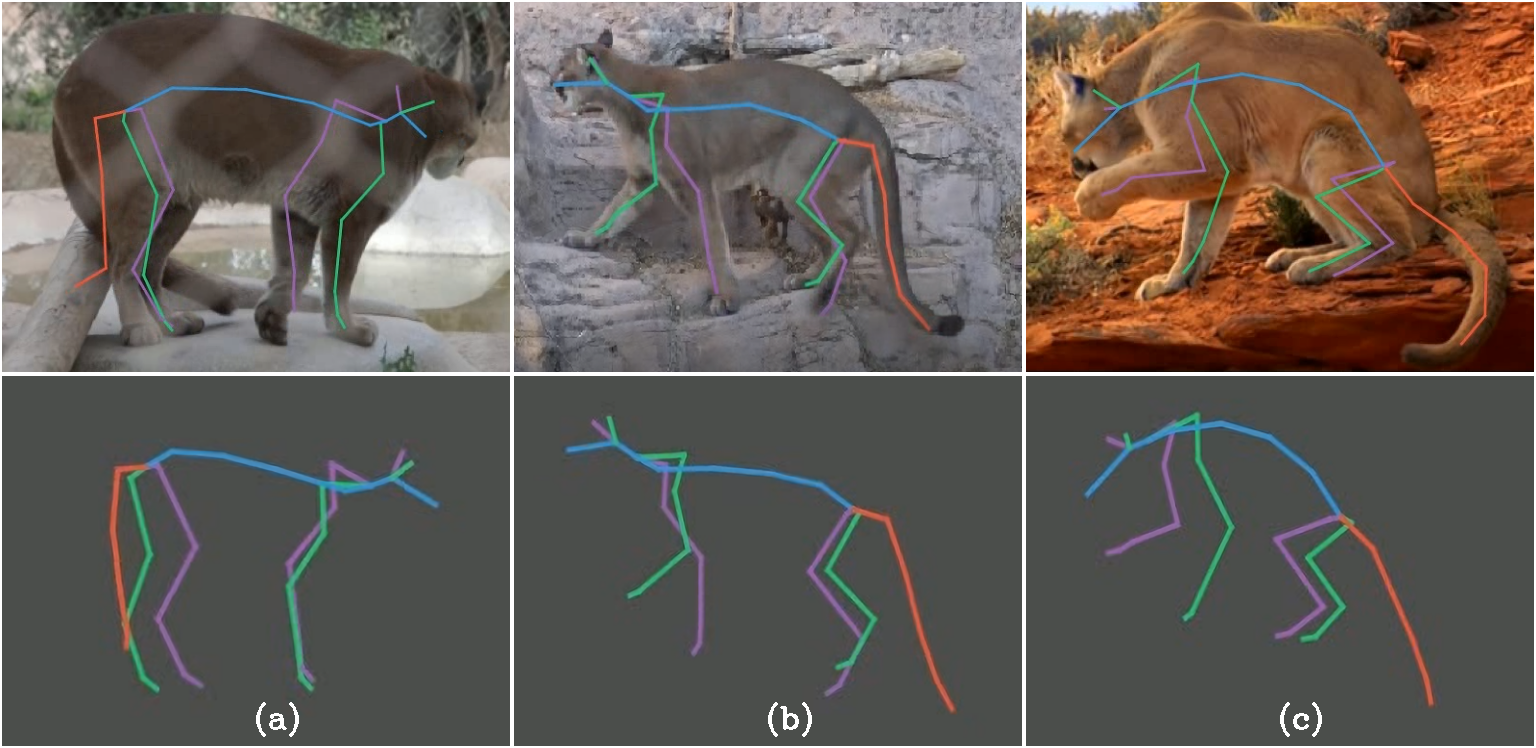}
  \caption{\label{fig:zbPipelineResults}
  Examples of 2D (above) and 3D (below) results on test data from the wild
            }
    \vspace{-4mm}
\end{figure}

\section{Conclusion}
Quadrupeds pose estimation is a challenging task due to the lack of training data. In this work, we proposed a strategy to fill this gap by generating a synthetic dataset based on existing keyframe animations. Using the data generated, we successfully trained 2D and 3D pose estimation models, and integrated them into an end-to-end pipeline that infers 3D animations from videos of real cougars (Figure \ref{fig:zbPipelineResults}). The pipeline still shows limitations when exposed to complex videos including multiple subjects, major occlusions, low contrast, etc. providing opportunities for further improvements on both the dataset generation and ML models development. This work could benefit from the contributions of both academic and industry peers who focus on quadrupeds pose estimation and animations.

\bibliographystyle{eg-alpha-doi} 
\bibliography{zoobuilder}

\newcommand{\etalchar}[1]{$^{#1}$}
\begin{thebibliography}{\uppercase{CHS{\etalchar{*}}18}}

\bibitem[CHS{\etalchar{*}}18]{cao2018openpose}
\textsc{Cao Z., Hidalgo G., Simon T., Wei S.-E., Sheikh Y.}:
\newblock Openpose: realtime multi-person 2d pose estimation using part
  affinity fields.
\newblock \emph{arXiv preprint arXiv:1812.08008} (2018).

\bibitem[DP{\etalchar{*}}10]{dunn20103d}
\textsc{Dunn F., Parberry I., et~al.}:
\newblock \emph{3D math primer for graphics and game development}.
\newblock Jones \& Bartlett Publishers, 2010.

\bibitem[RF18]{redmon2018yolov3}
\textsc{Redmon J., Farhadi A.}:
\newblock Yolov3: An incremental improvement.
\newblock \emph{arXiv preprint arXiv:1804.02767} (2018).

\bibitem[RIHL18]{rayat2018exploiting}
\textsc{Rayat Imtiaz~Hossain M., Little J.~J.}:
\newblock Exploiting temporal information for 3d human pose estimation.
\newblock In \emph{Proceedings of the European Conference on Computer Vision
  (ECCV)} (2018), pp.~68--84.

\bibitem[ZKL{\etalchar{*}}16]{zhou2016places}
\textsc{Zhou B., Khosla A., Lapedriza A., Torralba A., Oliva A.}:
\newblock Places: An image database for deep scene understanding.
\newblock \emph{arXiv preprint arXiv:1610.02055} (2016).

\end{thebibliography}


\end{document}